\documentclass[sigconf]{acmart}

\usepackage{microtype}
\usepackage{times,latexsym}
\usepackage{url}
\usepackage[T1]{fontenc}
\usepackage{times}
\usepackage{soul}
\usepackage{url}
\usepackage{graphicx}
\usepackage{amsmath}
\usepackage{amsthm}
\usepackage{booktabs}
\usepackage{algorithm}
\usepackage{algorithmic}
\usepackage{array}
\usepackage{latexsym}
\usepackage{comment}
\usepackage{amsfonts}
\usepackage{subfigure}
\usepackage{multirow}
\usepackage{multicol}
\usepackage{xcolor}
\usepackage{upgreek}
\usepackage{natbib}
\usepackage{xspace}
\usepackage{adjustbox}
\usepackage{wrapfig}
\usepackage{bm}
\usepackage{amsmath, amsfonts}
\usepackage{color}
\usepackage{colortbl}
\usepackage{CJKutf8}
\usepackage{booktabs}
\usepackage{array} 
\usepackage{mwe}
\usepackage{titlesec}
\usepackage{enumitem}
\usepackage{microtype}
\usepackage{fancypar}
\usepackage[normalem]{ulem}

\newcolumntype{R}[1]{>{\raggedleft\arraybackslash}p{#1}}
\newcolumntype{L}[1]{>{\raggedright\arraybackslash}p{#1}}

\newcolumntype{C}[1]{>{\centering\arraybackslash}p{#1}}
\newcolumntype{A}[1]{>{\raggedleft\arraybackslash}p{#1}}

\newcommand{\ModelName}{\textsc{HashTation}}
\newcommand{\RCModule}{\textsc{TAM}}
\newcommand{\REModule}{\textsc{EAM}}

\def\EM{{\mathcal E}}

\def\GM{{\mathcal G}}

\def\MM{{\mathcal M}}
\def\NM{{\mathcal N}}

\AtBeginDocument{%
  \providecommand\BibTeX{{%
    \normalfont B\kern-0.5em{\scshape i\kern-0.25em b}\kern-0.8em\TeX}}}

\copyrightyear{2023}
\acmYear{2023}
\setcopyright{acmlicensed}\acmConference[WWW '23]{Proceedings of the ACM Web Conference 2023}{May 1--5, 2023}{Austin, TX, USA}
\acmBooktitle{Proceedings of the ACM Web Conference 2023 (WWW '23), May 1--5, 2023, Austin, TX, USA}
\acmPrice{15.00}
\acmDOI{10.1145/3543507.3583194}
\acmISBN{978-1-4503-9416-1/23/04}

\begin{document}

\title{Hashtag-Guided Low-Resource Tweet Classification}

\author{Shizhe Diao}
\authornote{Both authors contributed equally to this research.}
\affiliation{%
  \institution{The Hong Kong University of Science and Technology}
}
\email{sdiaoaa@connect.ust.hk}

\author{Sedrick Scott Keh}
\authornotemark[1]
\affiliation{%
  \institution{Carnegie Mellon University}
}
\email{skeh@cs.cmu.edu}

\author{Liangming Pan}
\affiliation{%
  \institution{University of California, Santa Barbara}
}
\email{liangmingpan@ucsb.edu}

\author{Zhiliang Tian}
\affiliation{%
 \institution{The Hong Kong University of Science and Technology}
}
\email{tianzhilianghit@gmail.com}

\author{Yan Song}
\affiliation{%
  \institution{University of Science and Technology of China}
}
\email{clksong@gmail.com}

\author{Tong Zhang}
\affiliation{%
 \institution{The Hong Kong University of Science and Technology}
}
\email{tongzhang@ust.hk}

\renewcommand{\shortauthors}{Diao and Keh, et al.}

\begin{abstract}
Social media classification tasks (\textit{e.g.}, tweet sentiment analysis, tweet stance detection) are challenging because social media posts are typically short, informal, and ambiguous. 
Thus, training on tweets is challenging and demands large-scale human-annotated labels, which are time-consuming and costly to obtain. 
In this paper, we find that providing hashtags to social media tweets can help alleviate this issue because hashtags can enrich short and ambiguous tweets in terms of various information, such as topic, sentiment, and stance. 
This motivates us to propose a novel \textbf{Hash}tag-guided \textbf{T}weet Classific\textbf{ation} model (\textbf{\ModelName}), which automatically generates meaningful hashtags for the input tweet to provide useful auxiliary signals for tweet classification. 
To generate high-quality and insightful hashtags, our hashtag generation model retrieves and encodes the post-level and entity-level information across the whole corpus. 
Experiments show that {\ModelName} achieves significant improvements on seven low-resource tweet classification tasks, in which only a limited amount of training data is provided, showing that automatically enriching tweets with model-generated hashtags could significantly reduce the demand for large-scale human-labeled data. 
Further analysis demonstrates that {\ModelName} is able to generate high-quality hashtags that are consistent with the tweets and their labels. The code is available at \url{https://github.com/shizhediao/HashTation}.
\end{abstract}





\maketitle

\section{Introduction}
\label{sec:intro}
\begin{table}[ht]
    \centering
    \caption{Examples of how hashtags can provide auxiliary information for better tweet classification.}
    \vspace{-0.3cm}
    \includegraphics[scale=0.33]{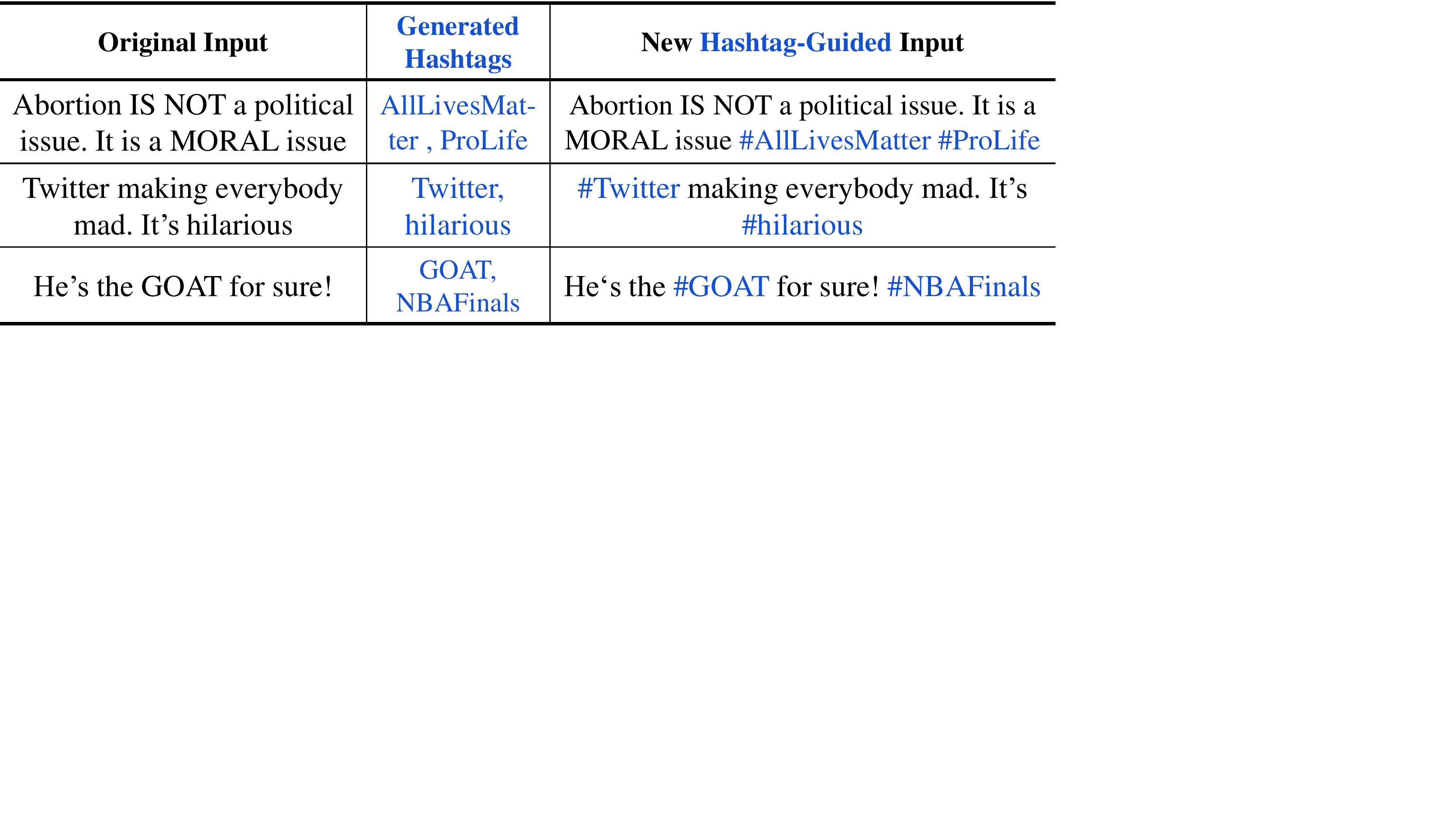}
    \label{fig:my_label}
    \vspace{-4.2cm}
\end{table}

Tweet Classification (TC) is an essential task in social media content analysis, which aims to analyze user behaviors and attitudes on Twitter. 
Typical tasks in this area include stance detection~\citep{mohammad2016semeval}, sentiment analysis~\citep{rosenthal2017semeval}, and
hate speech detection~\citep{basile-etal-2019-semeval}.
However, these tasks are often difficult because tweets are usually informal, idiosyncratic, and short in length and thus provide limited and ambiguous information.
Additional context and background knowledge are often needed to understand the content of a tweet better.
Due to this lack of information and the ambiguous nature of tweets, we often need to label a large-scale training corpus in order to train a satisfactory TC model~\citep{barbieri2018semeval,rosenthal2017semeval}.
However, the rapidly changing and evolving nature of social media content makes it challenging to annotate in-domain training data in a timely manner. 
Furthermore, data annotation is time-consuming and costly. 
To address the above challenges, we propose a model, {\ModelName}, with two novel features: 1) it can automatically enrich the content of social media tweets by \textit{hashtag generation}, and 2) the hashtag-enriched tweet classification model works well under the \textit{low-resource} setting in which only a limited amount of labeled data is available.

\begin{table*}[t]
    \centering
    \sc
    \caption{Pilot studies on TC.
    First, we split all the training data and validation data into two sets: tweets containing hashtags (\textit{w.} hashtags) and tweets without hashtags (\textit{w.o.} hashtags). We separately train and evaluate the model's performance on these two sets over 10 different random seeds.
    F1 scores on dev. set are reported.}
    \begin{tabular}{c|c|c|c|c|c|c|c|c}
        \toprule
         & \textbf{Emoji} & \textbf{Emotion} & \textbf{Hate} & \textbf{Irony} & \textbf{Offensive} & \textbf{Sentiment} & \textbf{Stance} & \textbf{Avg.} \\
         \midrule
         \textit{w.o.} Hashtags & 8.26 & 55.9 & 63.7 & 61.7 & 64.1 & 50.2 & 55.5 & 
         51.3 \\
         \textit{w.} Hashtags & 8.58 & 64.5 & 73.5 & 67.3 & 65.8 & 52.0 & 59.4 & 
         55.9 \\ 
         \bottomrule
    \end{tabular}
    \label{tab:pilot_study}
    \vspace{-0.2cm}
\end{table*}

Hashtags are commonly contained within tweets or appended to the end of tweets.
They not only facilitate rapid lookup for specific themes or web contents, but also contain important information that helps to enrich and disambiguate the contents of tweets. 
As exemplified in Table~\ref{fig:my_label}, we can hardly understand the topic and sentiment of the tweet ``Abortion IS NOT a political issue. It is a MORAL issue.'' without its hashtag ``\#RoeVWade''. 
Our pilot study~(Table~\ref{tab:pilot_study}) finds that hashtags provide important signals for tweet classification; the tweets without hashtags suffer from an average F1-score drop of 4.6\% in seven different tasks of TC compared with the tweets containing hashtags. 
Through this preliminary experiment, we find evidence to support our intuition that hashtags enrich the typically short and ambiguous tweets.

Unfortunately, despite the usefulness of hashtags, a vast majority of tweets do not use them (See Table \ref{tab:dataset_statistics}).
Motivated by this, our model {\ModelName} addresses low-resource tweet classification in two steps: 1) we propose a novel hashtag generation model by collecting hashtag-containing tweets as our training data, and 2) we enrich the hashtag-absent tweets using the generated hashtags from {\ModelName}. 
To generate more accurate and insightful hashtags, we leverage global contexts to consider not only the input tweet but also other relevant tweets, as well as tweets which share the same entities. 
To do this, we propose two corpus-level attention modules (Tweet Attention Module and Entity Attention Module) to retrieve helpful information across the whole corpus and entity graph. 
After obtaining generated hashtags, we compose a hashtag-prompted input by combining the hashtags and the original tweet with human-designed templates. 
We then feed the hashtag-prompted input into a pre-trained TC model.

Experiments on seven diverse TC tasks show that automatically enriching tweets with model-generated hashtags could significantly reduce the demand for large-scale human-labeled data.  
Further analysis shows that {\ModelName} generates high-quality hashtags that are consistent with the tweets and their labels, revealing the effectiveness of leveraging global contexts by retrieval. 

Our main contributions are as follows: 

\noindent $\bullet$ We empirically reveal that hashtags provide crucial signals for classifying social media posts in seven different tasks. 

\noindent $\bullet$ In light of the empirical evidence, we propose a novel hashtag generation model that leverages multi-grained global contexts to consider not only the input post but also the relevant posts and entities by using a cross-attention module and a graph entity network. 

\noindent $\bullet$ By enriching social media posts with generated hashtags, we significantly reduce the demand for human annotation for tweet classification. 
Our model significantly boosts low-resource tweet classification performance in seven different tasks.

\section{Approach}
\begin{figure*}[t]
\begin{center}
\includegraphics[scale=0.45, trim=0 0 170 0]{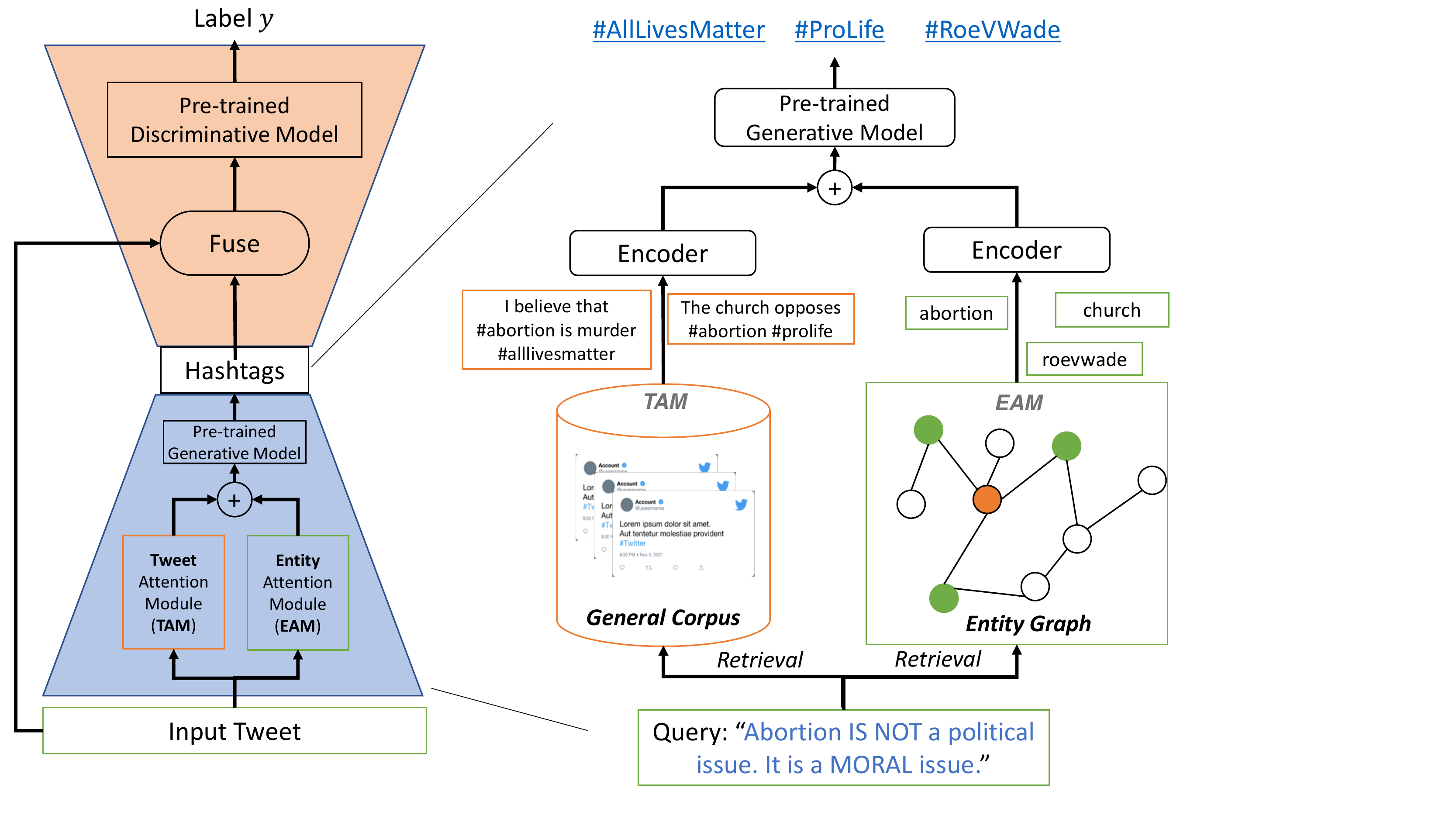}
\end{center}
\vspace{-0.4cm}
\caption{Left: Overview of {\ModelName}. 
There are two main components: the bottom module is the hashtag prompt generator and the top module is the classification model. 
Right: Illustration of tweet attention module (TAM) and entity attention module (EAM). 
The hashtag generator encodes the input tweets and decodes the hashtags via self-attention.
The TAM retrieves and incorporates the relevant semantic information at the tweet-level implicitly. 
The EAM retrieves and leverages the relevant semantic information at the entity-level explicitly.
The pre-trained discriminative model is adopted with a classification head to perform tweet classification.
}
\label{modelgraph2}
\end{figure*}

Figure \ref{modelgraph2} shows the architecture and detailed components of {\ModelName}. They are as follows: 
1) \textit{Hashtag Generator}, which encodes the input tweets and decodes the hashtags via self-attention; 
2) \textit{Tweet Attention Module} ({\RCModule}), employing a cross-document attention network to retrieve and incorporate the relevant semantic information at the tweet-level implicitly, which is then fused with the input representation;
3) \textit{Entity Attention Module} ({\REModule}), which employs a graph attention network to retrieve and leverage the relevant semantic information at the entity-level explicitly, which is then fused with the input representation;
4) \textit{Tweet Classifier}, which adopts a Transformer encoder with a classification head to perform tweet classification.
Below, we formulate the problem and then describe the details of each module.

\subsection{Problem Formulation}
In our setting, the basic input unit is a tweet $x = \{w_1 \cdots w_n\}$ where $w_i$ is the $i$-th word of the given tweet.
Output $H = \{ h_1 \cdots h_m \}$ is a sequence of all the hashtags of $x$, and $y$ is the target class/label corresponding to the input tweet. 
The first stage of our model, \textsc{Hash-Gen}, for hashtag generation can be formulated as follows. 
\begin{equation}
\setlength\abovedisplayskip{5pt}
\setlength\belowdisplayskip{5pt}
    H = \textsc{Hash-Gen}(x, {\RCModule}(x, \mathcal{D}), {\REModule}(x, \mathcal{D})),
\end{equation}
where \textsc{Hash-Gen} refers to the hashtag generator, {\RCModule} refers to the Tweet Attention Module that produces latent topic embeddings (Section \ref{subsec:tam}), and {\REModule} is the Entity Attention Module that produces explicit relevant entity information (Section \ref{subsec:eam}). 
Both TAM and EAM exploit a collection of tweets $\mathcal{D}$. 
The hashtag generation is then enhanced with the latent topics provided in $\mathcal{D}$ and the explicit entity relations provided in the {\REModule}.

The second stage of our model, \textsc{Tweet-Classifier}, for tweet classification is formulated as follows.
\begin{equation}
y = \textsc{Tweet-Classifier}(Fuse(H, x)),
\end{equation}
where \textsc{Tweet-Classifier} is a Transformer-based text classifier.
Details of the \textsc{Hash-Gen} and \textsc{Tweet-Classifier}, as well as how we integrate them, are described in the following subsections.

\subsection{Tweet Attention Module (TAM)}
\label{subsec:tam}
Given an input tweet $x$, relevant tweets usually share similar topics, which are good references to help determine what could be the optimal hashtags to describe $x$.
For example, for a tweet about school closures, the hashtag `\#\textit{CloseTheSchools}' may be absent in the tweet but may appear in other relevant tweets, providing helpful guidance for hashtag generation in this scenario.

To represent and exploit latent topics from relevant tweets,
we first aggregate all tweets from a collection (\textit{i.e.,} the union of both training and validation set), represented as $\mathcal{D} = \{ {x_1}, ..., {x_k}, ..., {x_l} \}$. Following the Transformer architecture, we define key vectors $\mathcal{U} = \{{\bm{u}_1}, ..., {\bm{u}_k}, ..., {\bm{u}_l} \}$ and value vectors $\mathcal{V} = \{{\bm{v}_1}, ..., {\bm{v}_k}, ..., {\bm{v}_l} \}$ with ${\bm{u}_k}$ and ${\bm{v}_k}$ corresponding to $x_k$.
Specifically,
${\bm{u}_k}$ is used to compute similarity with the input while ${\bm{v}_k}$ carries $x_k$'s encoding information for generating the final output, which acts as the latent topic embedding.
Then for each input $x$, we represent it through its sentential encoding $\bm{e}$ and use it as the `query' vector to address relevant tweets.
In detail, the addressing operation can be formalized as
\begin{equation}
    \label{eq:p}
    p_k = \frac{\exp(\bm{e}^\top \cdot
      {\bm{u}_k})}{\sum_{k=1}^{l}\exp(\bm{e}^\top \cdot
      {\bm{u}_k})} ,
\end{equation}
and for the entire set $\mathcal{D}$, we have
$
\bm{o} = \sum_{k=1}^l p_k {\bm{v}_k} ,$
where $\bm{o}$ is the output vector of the {\RCModule} to represent the latent topics from relevant tweets via a weighted encoding.

\begin{table*}[t]
    \small
    \centering
    \caption{Dataset statistics of the original dataset and the set whose tweets contain hashtags.}
    \vspace{-0.2 cm}
    \begin{tabular}{l|c|ccc|ccc}
        \toprule
        \multirow{2}{*}{\textbf{Task}} & \multirow{2}{*}{\textbf{Labels}} & \multicolumn{3}{c}{Original Dataset} & \multicolumn{3}{c}{With Hashtags} \\
         &  & $|$\textbf{Train}$|$ & $|$\textbf{Val}$|$ & $|$\textbf{Test}$|$ & $|$\textbf{Train}$|$ & $|$\textbf{Val}$|$ & $|$\textbf{Test}$|$ \\
        \midrule
        Emoji Prediction & 20 different emojis & 44489 & 4930 & 49664 & 19878 & 1800 & 20387 \\
        Emotion Recognition & anger, joy, sadness, optimism & 3188 & 363 & 1384 & 1394 & 178 & 650 \\
        Hate Speech Detection & hateful, not hateful & 8914 & 990 & 2963 & 2223 & 228 & 1394 \\
        Irony Detection & irony, not irony & 2802 & 942 & 781 & 1017 & 368 & 530 \\
        Offensive Language Identification & offensive, not offensive & 11616 & 1301 & 852 & 1664 & 194 & 630 \\
        Sentiment Analysis & positive, neutral, negative & 45612 & 2000 & 12200 & 8262 & 347 & 4571 \\
        Stance Detection & in favor, neutral, against & 2620 & 294 & 1249 & 2578 & 288 & 1249 \\
        \bottomrule
    \end{tabular}
    \label{tab:dataset_statistics}
\end{table*}

\subsection{Entity Attention Module (EAM)}
\label{subsec:eam}
\textbf{Graph Construction.}
To better aggregate and leverage the entity information across all tweets, the entity attention module offers strong clues in determining what to predict next.
We use a named entity recognition model to construct the entity graph by extracting relevant entities across all tweets based on co-occurrence. 
We first create an empty graph $\GM = (\NM, \EM)$, where $\NM = \{n_1, \cdots, n_i, \cdots, n_N\}$ and $\EM = \{e_1, \cdots, e_j, \cdots, e_M\}$ are the node and edge sets, respectively.
For each tweet input $x$, we use our entity recognizer to extract a set of entities.
Each entity will be added into $\GM$ as a node and linked to those nodes that are extracted from the same tweet.
As a result, we obtain an entity graph with multiple nodes and edges, capturing the critical entity relations across relevant tweets. 

\textbf{Graph Encoding.}
The node embeddings are randomly initialized as $\bm{H}^{(0)}$ during graph construction.
In order to equip them with better semantic relation information, we adopt the graph convolutional networks~\citep{KipfW17} to provide a better initialization.
We then adopt the idea of graph attention networks (GAT)~\citep{GAT2018graph} to obtain the node encoding by aggregating information from their neighbors and update it with dynamic weights.
Formally, GAT accepts $\bm{H}^{(0)}$ and outputs $\bm{H}^{(L)}$ after conducting $L$ layers of state transitions.
Given a sequence of hidden representations $\bm{H}^{(l)} = [{\bm{\vec{h}}_1^l, \cdots, \bm{\vec{h}}_i^l, \cdots, \bm{\vec{h}}_N^l}]$ at layer $l$ with $\bm{\vec{h}}_i^l$ indicating the feature representation for $i$-th node,
and an adjacency matrix $A$,
the hidden representation at layer $l+1$ is calculated by
\begin{equation}
    \bm{H}^{(l+1)} = \sigma\left(\bm{\hat{D}}^{(-\frac{1}{2})}\bm{\hat{A}}\bm{\hat{D}}^{-\frac{1}{2}}\bm{H}^{(l)}\bm{W}^{(l)}\right),
\end{equation}
where $\bm{\hat{A}} = \bm{A} + \bm{I}$, $\bm{I}$ is the identity matrix and $\bm{\hat{D}}$ is the diagonal node degree matrix of $\bm{\hat{A}}$.
$\bm{W}^{(l)}$ is a weight matrix for the $l$-th neural network layer, and $\sigma(\cdot)$ is the $ReLU$ activation function.
Then, we calculate the attention coefficient between nodes $n_i$ and $n_j$ by
\begin{equation}
    \alpha_{ij} = \frac{\exp \left([\bm{W}^Q\bm{\vec{h}}_i^l \bm{W}^K \bm{\vec{h}}_j^l]\right)}{\sum_{k \in \MM_i} \exp\left([\bm{W}^Q\bm{\vec{h}}_i^l \bm{W}^K \bm{\vec{h}}_k^l]\right)},
\end{equation}
where $\bm{W}^Q$ and $\bm{W}^K$ are weight matrices for feature transformation of query and key, and $\MM_i$ is the first-order neighbors of node $n_i$ (including $n_i$).

Third, after obtaining attention coefficients, the output encoding is a linear combination of the features in its neighbors, computed by
\begin{equation}
\label{eq:4}
    \bm{\vec{h}}^{l+1}_i = \sigma \left(\sum_{n_j \in \MM_i} \alpha_{ij} \bm{W} \bm{\vec{h}}^{l}_j \right),
\end{equation}
where $\bm{W}$ is a weight matrix and $\sigma(\cdot)$ is the $ReLU$ activation function.

\subsection{Integrating Tweet-level and Entity-level Information} 

Although RNN-based sequence-to-sequence models are widely used for the hashtag generation task,
we use the Transformer \citep{vaswani2017attention} as the backbone encoder-decoder framework in this paper. 
This has been proven to be more effective than RNN-based sequence-to-sequence models in many generation tasks \citep{vaswani2017attention, keskar2019ctrl, khandelwal2019sample, hoang2019efficient, liu2019hierarchical}.
Once the latent topic embedding $\bm{o}$ and graph entity encoding $\bm{\vec{h}}$ are obtained, we combine them with the Transformer encoding-decoding process via the following steps.

First, the tweet input is passed through the Transformer encoder, which results in a hidden state $\bm{h}_i$ for each input token. 
Then we combine $\bm{h}_i$ and $\bm{o}$ via element-wise addition $\bm{\tilde{h}}_i = \bm{h}_i + \bm{o}$.
Third, we enhance $\bm{\tilde{h}}_i$ with graph entity information $\bm{\vec{h}}$ by
$
    \bm{\tilde{h}}_i = \bm{\tilde{h}}_i + \sum_{j}\bm{\vec{h}}_{i,j},
$
where $\bm{\vec{h}}_{i,j}$ is the graph encoding of the $j$-th entity node associated to the $i$-th token.
$\bm{\tilde{h}}_i$ is the final encoding state of $i$-th token
and is sent to the decoding process through each multi-head attention layer to calculate the attention vector $\bm{a}^t = {\alpha^t_1} ...{\alpha^t_i}...{\alpha^t_n}$ at each decoding step $t$.
Next, $\bm{a}^t$ is used to produce the context vector $\bm{c}^t = \sum_{i=1}^n \alpha^t_i \bm{{\tilde{h}}}_i$
by a weighted sum of the encoding hidden states.
Later $\bm{c}^t$ is concatenated with the decoder output $\bm{s}^t$ and then fed into a single linear layer, followed by a softmax function, to produce the vocabulary distribution for the output word at step $t$
\begin{equation}
\label{p_vocab}
\bm{d}_{v} =
\frac{\exp(\bm{z}^t)}{\sum\limits_{V}\exp(\bm{z}^t)} ,
\end{equation}
where $\bm{z}^t = \bm{W_{1}}(\bm{W_{2}} \cdot (\bm{s}^{t} \oplus \bm{c}^t))$, a vector with $|V|$ dimension and $V$ is the predefined vocabulary providing word candidates for hashtag generation.
$\bm{W_{1}}$ and $\bm{W_{2}}$ are trainable parameters for the two aforementioned linear layers, respectively.

\subsection{Tweet Classifier}
After obtaining the desired hashtags,  we fuse the hashtags $H = h_{1}, h_{2}, .., h_{n}$ with the input tweet $x$ to obtain the hashtag-guided tweet $Fuse(H, x)$. 
There are several ways to implement the $Fuse$ function like simple concatenation, prompting with a manual template, and so on. 
We explore the effects of different strategies in Section~\ref{sec:effects_fusion}.
Finally, we train a tweet classifier $\mathcal{F}$ with the following objective:
\begin{equation}
\mathop{min}\limits_{\phi}(\mathcal{L}(\mathcal{F}(Fuse(H, x)), Y)),
\label{eq:fuse}
 \end{equation}
where $\mathcal{L}$ is the cross-entropy loss function following the standard practice in tweet classification~\citep{loureiro-etal-2022-timelms,nguyen2020bertweet}. 

\section{Experimental Settings}
In this section, we first introduce the datasets (Section \ref{sec: Datasets}), followed by the baseline models (Section \ref{sec: Baselines}) and evaluation metrics (Section \ref{sec: Evaluation}).  
Lastly, we describe the implementation details (Section \ref{sec: Implementation}) for various experiments.

\subsection{Datasets}
\label{sec: Datasets}
We perform our experiments on 7 diverse tweet classification tasks in the TweetEval benchmark datasets~\citep{barbieri-etal-2020-tweeteval}. 
All 7 tasks are taken from previous SemEval tasks (and corresponding datasets) and are as follows: emotion recognition~\citep{mohammad2018semeval}, emoji prediction \citep{barbieri2018semeval}, irony detection~\citep{van2018semeval}, hate speech detection~\citep{basile-etal-2019-semeval}, offensive language identification~\citep{zampieri2019semeval}, sentiment analysis~\citep{rosenthal2017semeval}, and stance detection~\citep{mohammad2016semeval}. Additional details about these tasks can be found in Table~\ref{tab:dataset_statistics}.

\begin{table*}[t]
    \small
    \centering
    \sc
    \caption{F1-scores for baseline models (top) and our proposed method and its variants (bottom). 
    Best results are highlighted in \textbf{bold}.
    We report the average score over ten different random seeds.
    $^{^{*}}$ denotes {\ModelName} has significant differences ($p-value$ \textless  0.05) over baseline models.
    }
    \vspace{-0.2cm}
    \begin{tabular}{c|cccccccc}
        \toprule
         & \textbf{Emoji} & \textbf{Emotion} & \textbf{Hate} & \textbf{Irony} & \textbf{Offensive} & \textbf{Sentiment} & \textbf{Stance} & \textbf{Avg.}\\
        \midrule
        Kim-CNN & 3.8 & 20.1 & 51.1 & 39.8 & 41.9 & 39.5 & 32.2 & 32.6 \\
        BiLSTM & 5.9 & 26.8 & 49.2 & 37.4 & 47.6 & 43.1 & 28.7 & 34.1 \\
        BERT-base & 11.2 & 57.7 & 51.6 & 53.7  & 56.4  & 56.6  & 52.3  & 
        48.5 \\
        RoBERTa-base & 11.8  & 58.9  & 56.7  & 54.2  & 59.7  & 57.3 & 54.1  & 
        50.4\\
        \hline
        BERTweet & 12.1  & 59.4  & 55.5  & 57.0  & 61.8 & 59.0  & 55.5  &  51.5 \\
        {\ModelName}-BT & 12.7 & 61.0 & 56.4 & 58.6 & \textbf{63.8$^{^{*}}$}  & \textbf{59.9$^{^{*}}$} & 56.3 & 
        52.7\\
        
        w/o TAM & 12.5 & 60.6 & 56.3 & 57.7& 63.0 & 59.8 & 56.2 & 
        52.3 \\
        
        w/o EAM & 12.6 & 60.5 & 55.7 & 58.6 & 63.5 & 59.4 & 56.3 & 
        52.4 \\
        
        \hline
        TimeLMs & 12.4 & 60.2 & 56.9 & 59.6 & 60.0  & 57.4 & 55.9 & 51.8\\
        {\ModelName}-TL & \textbf{13.0$^{^{*}}$}  & \textbf{61.6$^{^{*}}$} & \textbf{58.0$^{^{*}}$} & \textbf{60.9$^{^{*}}$} & 
        62.3 & 59.6 & \textbf{56.5$^{^{*}}$} & 
        \textbf{53.1$^{^{*}}$} \\
        w/o TAM & 12.7 & 61.2 & 57.6 & 60.6 & 61.9  & 58.9 & 56.1  & 
        52.7 \\
        w/o EAM & 12.7 & 61.3  & 57.6 & 60.7 & 61.6 & 59.0 & 56.2 & 
        52.7 \\
        \bottomrule
    \end{tabular}
    \vspace{-0.2cm}
    \label{tab:main_results1}
\end{table*}

\textbf{Creating the hashtag generation input and output.} For hashtag generation, we train a separate hashtag generator for each task, as the datasets cover significantly different domains. 
To extract hashtags, we searched for appearances of the octothorpe symbol ($\#$) and considered the contiguous string following it, until a whitespace is reached. These hashtags can either appear mid-tweet or at the end of the tweet. For hashtags that appear mid-tweet, we remove the hashtag symbol ($\#$) but keep the word itself, as removing these words may disrupt the coherence of the sentence. On the other hand, for hashtags that appear at the end of the tweet, we completely remove these from the body of the tweet. If we keep these words at the end of the sentence, the hashtag generator will easily pick up on these false signals and just learn to return the last few words of every sentence without actually learning to perform hashtag generation. By considering this setting of keeping in-sentence hashtags and removing end-of-sentence hashtags, we are able to train the model to perform prediction for both present hashtags and absent hashtags. 
Examples of this overall process are outlined in Table \ref{tab:hashtag-dataset}.
Similar to \citet{wang-etal-2019-microblog}, we implement some preprocessing steps to clean up our tweets. 
Links, mentions, and numbers were replaced with ``URL'', ``MENTION'', and ``DIGIT'', respectively. 
The details of the processed dataset are shown in Table \ref{tab:dataset_statistics}.

\textbf{Low-resource setting.} 
As discussed in Section \ref{sec:intro}, real-life classification labels are often difficult to acquire for the rapidly changing landscape of tweets.
As such, we conduct our experiments in a low-resource setting, where we randomly select a subset from the training set. 
To keep the sizes of different datasets relatively consistent, we use the following sampling ratios: If the size of the original dataset is $<5000$, we sample $10\%$ of the dataset. 
If it's between $5000$ and $10000$, we sample $5\%$, and if it's $>10000$, we sample $1\%$.
All experiments are performed over ten different random seeds. 

\subsection{Baselines}
\label{sec: Baselines}
To verify the effectiveness of our {\ModelName} model, we compare its performance against the following baseline classification models: \textbf{Kim-CNN} \cite{kim-2014-convolutional}, \textbf{BiLSTM} \cite{bilstm-1997}, \textbf{BERT}~\citep{devlin2018bert}, \textbf{RoBERTa} \cite{Liu2019RoBERTaAR}, \textbf{BERTweet}~\citep{nguyen-etal-2020-bertweet}, and \textbf{TimeLMs}~\citep{loureiro-etal-2022-timelms}.
BERTweet and TimeLMs are two state-of-the-art models that are pre-trained on 850 million and 124 million English tweets, respectively.

\begin{itemize}[leftmargin=*,label=$\bullet$,noitemsep,partopsep=0pt,topsep=0pt,parsep=0pt]
    \item \textbf{Kim-CNN} \cite{kim-2014-convolutional}. A simple convolutional neural networks framework with a classification layer attached to the end. For this CNN-based model, we use kernel sizes of 2,3,4,5 and 64 filters for each, with a dropout of 0.5 before the linear layer.
    \item \textbf{BiLSTM} \cite{bilstm-1997}. 
    A bidirectional long short-term memory network considering the temporal order of words in the tweet. The hidden size and the dropout rate for the LSTM are set to 512 and 0.2, respectively. We use an LSTM hidden size of 512 and an LSTM dropout of 0.2. In addition, we use a dropout of 0.5 and a size of 32 for the final linear classification layer.
    \item \textbf{BERT}~\citep{devlin2018bert}. A BERT-base model pre-trained on generic corpus with a classification head fine-tuned on specific task.
    \item \textbf{RoBERTa} \cite{Liu2019RoBERTaAR}. 
    A RoBERTa-base model pre-trained on generic corpus with a classification head fine-tuned on specific task. 
    \item \textbf{BERTweet}~\citep{nguyen-etal-2020-bertweet}. A RoBERTa-base model pre-trained on 850M English Tweets with a classification head fine-tuned on specific task.
    \item \textbf{TimeLMs}~\citep{loureiro-etal-2022-timelms}. A RoBERTa-base model specialized on diachronic Twitter data and pre-trained on 124M English Tweets.
    \item \textbf{{\ModelName}} (and its variants) -- For the \textsc{Hash-Gen} hashtag generator, we use the Huggingface BART-base model, while for the tweet classifier, we use the BERTweet and TimeLMs. Additionally, for the document embeddings used in the Tweet Attention Module (TAM), we used the \texttt{all-MiniLM-L6-v2} model from the sentence-transformers \footnote{\url{https://www.sbert.net/}} library.
\end{itemize}

\subsection{Evaluation Metrics}
\label{sec: Evaluation}
For evaluation, we use the respective metrics
following TweetEval benchmark~\citep{barbieri-etal-2020-tweeteval}, which are detailed as follows.
For \textsc{emoji}, \textsc{emotion}, \textsc{hate}, and \textsc{offensive}, we use the macro F1-score. 
For \textsc{irony}, we use the F1-score on the positive class. For \textsc{sentiment}, we use the macro recall score. 
Lastly, for \textsc{stance}, we use the average F1 scores of the ``against'' class and the ``favor'' class.

\subsection{Implementation}
\label{sec: Implementation}
For {\RCModule}, we utilize sentence-transformer~\citep{reimers-2019-sentence-bert} to initialize key vectors $\bm{u}_k$ and value vectors $\bm{v}_k$ to guarantee reliable addressing as a warm start for those vectors and they are updated during the training process.
Different from $\bm{u}_k$ and $\bm{v}_k$, the sentential encoding $\bm{e}$ of each input tweet $x$ is represented as the average of its word representations which are randomly initialized to ensure their compatibility with the backbone Transformer's vector space during training.
For {\REModule}, we extract relevant entities with social media NER systems\footnote{\url{https://github.com/napsternxg/TwitterNER}} to construct the graph from the corpus.
We use the AdamW~\citep{loshchilov2018decoupled} optimizer, together with a batch size of 16. 
We pad or truncate the inputs so that all of them have a length of 64. 
For the CNN and LSTM-based methods, we use a learning rate of 2e-5, while for our {\ModelName} model, we use a learning rate of 1e-5 for both the hashtag generator and the tweet classifier. 
Beam search is applied to generate multiple phrases during hashtag generation with a beam size of 10 and a maximum sequence length of 40.
These experiments were performed on Nvidia 2080Ti GPUs with 11 GB memory.

\section{Experimental Results}
\begin{table*}[t]
    \small
    \centering
    \sc
    \caption{Effects of different fusion methods on the TimeLMS model. 
    We explore four methods: no hashtags, standard, pre-pending at the start, appending at the end.
    F1-scores are reported on seven tasks.}
    \begin{tabular}{c|ccccccc}
    \toprule
        \textbf{Fusion Method} & Emoji & Emotion & Hate & Irony & Offensive & Sentiment & Stance \\
        \midrule
        Without Hashtags & 12.4 & 60.2 & 56.9 & 59.6 & 60.0 & 57.4 & 55.9 \\
        Standard & \textbf{13.0} & 61.6  & \textbf{58.0} & \textbf{60.9} & \textbf{62.3} & \textbf{59.6} & 56.5 \\
        Start & 12.9 & 61.0 & \textbf{58.0} & 60.7 & 61.3 & 59.3 & 56.1 \\
        End & \textbf{13.0} & \textbf{61.7} & 57.6 & 60.5 & 61.9 & 59.6 & \textbf{56.6} \\
    \bottomrule
    \end{tabular}
    \label{tab:fusion_methods}
    \vspace{-0.2cm}
\end{table*}

The results on seven benchmark datasets are reported in Table~\ref{tab:main_results1}.
Overall, we observe that our {\ModelName} model performs the best, outperforming both BERT-base and RoBERTa-base by a significant margin, as well as displaying gains of 1.2\% and 1.3\% over the state-of-the-art BERTweet and TimeLMs model, respectively. 
This shows that the hashtag generator and tweet classifier indeed help the model to make the right predictions.
The first two models (Kim-CNN and BiLSTM) are not large language models, and they perform poorly on this low-resource classification task, with an average performance of 32.6 and 34.1, respectively.
The first major performance improvement comes when we introduce the BERT and RoBERTa models, increasing the performance from 48.5 to around 50.4. 
This is due to the knowledge gained by the language model pre-training, which is helpful in tweet understanding, and consistent with previous language model literature~\citep{devlin2018bert, liu2019roberta}. 
Next, another significant jump in performance occurs between the RoBERTa-base model and domain-specific models (\textit{i.e.,} BERTweet and TimeLMs). 
This increase can be attributed to the additional pre-training in a specialized domain (social media texts like tweets).
This method is called domain-adaptive continual pre-training and has been verified in other domains, like biomedical domain~\citep{lee2020biobert}, clinical domain~\citep{alsentzer2019publicly}, scientific domain~\citep{beltagy2019scibert}, social media domain~\citep{loureiro-etal-2022-timelms,nguyen-etal-2020-bertweet} and so on.
The final major improvement in performance is from domain-specific pre-trained models to our {\ModelName} model. 
The main differences are the addition of our proposed modules, the {\RCModule} and the {\REModule}, as well as the incorporation of the generated hashtags.
All of these components complement each other to contribute to the performance improvement
\label{sec:ablation}
We also ablate certain components of our model to verify the contributions in Table \ref{tab:main_results1}.
For {\ModelName}-TL model, removing the {\RCModule} and {\REModule} causes an average of 0.4\% performance decrease equally.
We observed similar trends on the {\ModelName}-BT model as well.
These ablation studies confirm the motivation behind generating hashtags by retrieving global information carried by general corpus and entities.

\section{Analysis}
In this section, we investigate the effects of different fusion methods (Section ~\ref{sec:effects_fusion}), the performance on different datasets (Section ~\ref{sec:dataset-performance}), the quality of the generated hashtags (Section ~\ref{sec:hashtag-qual}), and we present a case study on specific examples (Section ~\ref{sec:case-study}).

\subsection{Effects of Fusion Methods}
\label{sec:effects_fusion}
There are several different ways to implement the $Fuse$ function mentioned in Equation~\ref{eq:fuse}.
To verify the effects of different fusion methods, we examine the following four templates: 
\begin{itemize}[leftmargin=*,label=$\bullet$,noitemsep,partopsep=0pt,topsep=0pt,parsep=0pt]
\item \textbf{Without Hashtags:} $Fuse(H, x) = x$
\item \textbf{Standard:} 
For each hashtag, we check whether it is a present or absent hashtag. For present hashtags, we prepend the hashtag symbol (\#) to the existing word corresponding to the hashtag; for absent hashtags, we append them to the end of the tweet.
\item \textbf{Start:} $Fuse(H, x) = [H, x]$
\item \textbf{End:} $Fuse(H, x) = [x, H]$
\end{itemize}
Here, $[\cdot, \cdot]$ represents the concatenation operation.
The performance on seven datasets is reported in Table~\ref{tab:fusion_methods}.
We find that there is no big difference between \textbf{Standard} and \textbf{End} while a slight performance drop is observed on \textbf{Start}. This is consistent with the general structure of tweets, as hashtags usually appear at the end of tweets rather than at the beginning. Nevertheless, all three with-hashtag methods still outperform the without-hashtag fusion, indicating that the mere inclusion of these guiding hashtags is sufficient to increase the performance of the model.

\subsection{Performance Across Datasets}
\label{sec:dataset-performance}
Across all 7 datasets, adding the hashtags boosts the performance, as seen in Tables \ref{tab:main_results1} and \ref{tab:fusion_methods}. However, improvement varies across datasets. For instance, with datasets such as \textsc{Irony} and \textsc{Emotion}, the improvement of adding hashtags (\textsc{BERTweet} vs \textsc{HashTation-BT} and \textsc{TimeLMs} vs \textsc{HashTation-TL}) is around 2\%, which is much higher than other datasets such as \textsc{Hate} or \textsc{Stance}. This tells us that certain types of data will benefit more from the enriched context than others. 
For instance, with \textsc{Emotion}, if the hashtag generation model is able to generate a hashtag such as ``happy'' or ``sad'', then they will serve as very strong signals for the classification model. 
In contrast, for a dataset such as \textsc{Hate}, the hashtag generation model might predict very useful hashtags in terms of topic (e.g. ``elections'', ``Ukraine''), but these might not be that useful for our particular downstream task (hate speech vs not hate speech) because both positive and negative classes will share these same topics.

\begin{table*}
\centering
\small
\addtolength{\tabcolsep}{-4pt}
\caption{Examples of hashtags generated by {\ModelName}, as compared to ground truth hashtags.}
\resizebox{\textwidth}{!}{
\begin{tabular}{|p{225pt}|p{45pt}|p{55pt}|p{130pt}|p{90pt}|}
\hline 
\textbf{Text} & \textbf{Dataset} & \textbf{Classification} & \textbf{Ground Truth Hashtags} & \textbf{Generated Hashtags} \\ \hline
\hline
Riding with @user and some incredible people. Truly magical. & Emoji & sparkles & \#soulcycle \#lululemon  & \#soulcycle \\
\hline
Oh Canada shouldn't be sung like that. & Emotion & Anger & \#terrible \#MLBTHESHOW17 & \#terrible \#Canada \\
\hline
Both \#Frightening and \#Demented \#Sick. Now it will be our problem. & Hate & No & \#Frightening, \#Demented, \#Sick, \#SendThemBack, \#KAG & \#Sick \#BuildThatWall \#MAGA \#KAG \\
\hline
Oh how I love working in Baltimore & Irony & Yes & \#not & \#not \#Baltimore \\
\hline
@user Ya  Obama on trade  2-YEARS AGO:  ``Trump is just NOT TELLING THE TRUTH  How STUPID could our leaders be''  God bless Trump! & Offensive & Yes & \#MAGA \#KAG & \#MAGA \#Trump \\
\hline
Messi: ``To have a good team, everything starts there'' & Sentiment & Joy & \#fcblive & \#fcblive \\
\hline
``Manspreading''? But women hog subway space, too! & Stance (Feminism) & against & \#doublestandards & \#doublestandards \#manspreading \\
\hline
\end{tabular}
}
\label{tab:hashtag-case}
\end{table*}

\subsection{Hashtag Quality}
\label{sec:hashtag-qual}
In our work, tweet classification is the downstream task, while hashtag generation is an important intermediate task. 
In this section, we check the quality of generated hashtags by comparing {\ModelName} with three state-of-the-art hashtag generation baselines (Bi-Attn~\citep{wang2019microblog}, One2Set~\citep{ye2021one2set}, and    SEGTRM~\citep{mao2022attend}).
To measure the quality of the generated hashtags, we adopt the precision, recall, and F1 metrics used in document retrieval. 
Additionally, we make a distinction between "present hashtags" which appear verbatim in the original tweet, and "absent hashtags" which do not match any contiguous word sequence in the original tweet. 
Results are reported in Table \ref{fig:hashtag-qual-avg}.
From the table, we conclude that {\ModelName} outperforms all baseline hashtag generation models which are specifically designed for this task, illustrating its effectiveness of leveraging tweet-level and entity-level information by TAM and EAM, respectively. 
Moreover, it is observed that absent hashtags are clearly much harder to predict than present hashtags, as evidenced by the difference in F1-scores (0.381 for present vs 0.122 for absent).
This is consistent with our expectation that it is a lot easier to generate hashtags by simply copying important words from the text as opposed to having to find a new (absent) word to use. 
Given this challenge, it is thus quite impressive that {\ModelName} is still able to correctly generate absent hashtags some of the time, considering that most of the input texts are very short. 
As observed in Table \ref{tab:hashtag-case}, there are certain input tweets from our dataset that are extremely short and seemingly contain no substantial content. 
For instance, for the tweet "Oh how I love working in Baltimore," {\ModelName} is able to connect it with the hashtag \#not, even though nowhere in the sentence is the word "not" mentioned. 
The reason {\ModelName} is able to predict this is because there are many other tweets in the corpus which are ironic and also share phrases like "how I love" in them. 
This is a testament to the effectiveness of our {\RCModule} and {\REModule} modules, which allow the {\ModelName} model to focus not only on the current text but also on several other relevant texts which may contain more useful information.
From Table \ref{fig:hashtag-qual-avg}, another key observation is that our model's precision scores are generally much higher than the model's recall scores. This means that when the model generates a hashtag, its prediction is likely to be correct, but there may be some ground-truth hashtags that the model fails to generate, which suggests that our model is generally quite conservative in making its predictions.

\begin{table}[t]
    \centering
    \caption{Average precision, recall, and F1 scores for the hashtag generation task across all datasets.}
    \begin{tabular}{c|ccc}
    \toprule
        Type & Precision & Recall & F1-Score \\
        \hline
        Bi-Attn~\citep{wang2019microblog} & 0.202 & 0.069 & 0.103 \\
        One2Set~\citep{ye2021one2set} & 0.231 & 0.095 & 0.134 \\
        SEGTRM~\citep{mao2022attend} & \textbf{0.352} & 0.127 & 0.187 \\
        {\ModelName} (ours) & 0.340 & \textbf{0.135} & \textbf{0.193} \\
        \hline
        Present & 0.848 & 0.234 & 0.381 \\
        Absent & 0.171 & 0.108 & 0.122 \\
    \bottomrule
    \end{tabular}
    \label{fig:hashtag-qual-avg}
    \vskip -1.5em
\end{table}

\begin{table*}[htbp]
\small
\centering 
\caption{An example of an input text whose stance is not very obvious and is implied instead of explicitly stated. We highlight the corresponding entities and relevant tweets that are extracted by the {\REModule} and {\RCModule} modules.}
\vspace{-0.3 cm}
\subtable[Example Input Text and Corresponding Hashtags (Entities from {\REModule} highlighted in \textcolor{blue}{blue})]{
\begin{tabular}{| p{10.0cm} | p{2.5cm} | p{2.5cm} | }
    \hline
    \textbf{Input Text} & \textbf{G. Truth Hashtags} & \textbf{Pred. Hashtags} \\
    \hline
    3 cases of COVID-19 (coronavirus) in 2 schools in my \textcolor{blue}{city}, both \textcolor{blue}{involving teachers coming} back from the \textcolor{blue}{north of Italy} and \textcolor{blue}{having contact} with the \textcolor{blue}{children} for almost a week before anything was done. & \#coronavirusuk, \#CloseTheSchools & \#CloseTheSchools \\
    \hline
\end{tabular}
}
\subtable[Tweets with Relevant Entities (extracted by {\REModule})] {
\begin{tabular}{| p{12.5cm} | p{3cm} |}
    \hline
    What a lack of intestinal fortitude MENTION you will have the deaths of Australians on your hands. When we reach 12 pages of obituaries like \textcolor{blue}{Italy} please look in the mirror for the cause. & \#lockusdown,  \#coronavirus, \#CloseTheSchools \\
    \hline
    At the end of the day, schools with sealed windows, and interior classrooms will have Coronavirus buildup that will increase COVID19 viral load! These building shouldn't be used. \textcolor{blue}{Children} and \textcolor{blue}{teachers} are not going to be victims of "risk mitigation" & \#Coronavirus, \#COVID19, \#NotMyChild \\
    \hline
\end{tabular}
}
\subtable[Most Relevant Tweets (extracted and ranked by {\RCModule})]{
\begin{tabular}{| p{12.3cm} | p{3cm} | }
    \hline
    We would all feel very different about schools reopening if we had a goverment that: -Was trustworthy -Had best interest of children as only motivation  -Based on Science  -provided necessary resources to ALL schools -had a National plan   None of these is true & \#NotMyChild \\
    \hline
    Schools should just be closed,the 2020 curriculum will continue once the vaccine is found, even if it means 2020 curriculum get done in 2021-2022, we’ll have to adjust and catch-up at some point. & \#SchoolsMustShutDown \\
    \hline
\end{tabular}
}
\label{tab:case-study}
\end{table*}

\subsection{Case Study}
\label{sec:case-study}
In Table \ref{tab:case-study}, we consider a stance detection task with the following example sentence: \begin{small} \textit{``3 cases of COVID-19 (coronavirus) in 2 schools in my city both involving teachers coming back from the north of Italy and having contact with the children for almost a week before anything was done.''} \end{small} This sentence has a stance in favor of school closures but is not very explicit in its claims. In fact, nowhere in the tweet are the words "school" or "closure" ever mentioned, nor does it use very pointed positive/negative words. Without the relevant surrounding context, it would be difficult for a model to identify this stance that is subtly hidden in a seemingly-neutral sentence.

With the {\REModule} and {\RCModule} modules, our {\ModelName} model is able to identify some relevant tweets and entities which explicitly discuss school closures and even have hashtags such as \#CloseTheSchools and \#SchoolsMustShutDown (see Table \ref{tab:case-study}). By identifying these relevant tweets, our model is able to gain a much broader context, providing key information beyond simply looking at the few words of the current tweet. Since our {\ModelName} model pays attention to these broader contexts instead of myopically focusing on just the current text, it becomes much easier to generate the correct hashtag, which greatly aids in stance classification.

\section{Related Work}
\subsection{Tweet Classification}
Tweet classification aims to classify a piece of the tweet into different categories, which is helpful for understanding public opinion.
Popular tweet classification tasks include stance detection~\citep{mohammad2016semeval}, sentiment analysis~\citep{rosenthal2017semeval}
emotion recognition~\citep{mohammad2018semeval}, hate speech detection~\citep{basile-etal-2019-semeval}, and so on.
Several studies explored SVM-based approaches \citep{dey2017twitter, sen2018stance}, CNN-based models \citep{vijayaraghavan-etal-2016-deepstance, wei-etal-2016-pkudblab} and RNN-based models \citep{zarrella-marsh-2016-mitre, siddiqua-etal-2019-tweet}, respectively.
Later, with the prevalence of Transformer architecture \citep{vaswani2017attention} and pre-trained language models \citep{devlin2018bert, Liu2019RoBERTaAR}, BERT-based approaches \citep{ghosh2019stance, li-caragea-2021-target} attracted a lot of attention and had proven their effectiveness in this task.
In addition to architectural improvement, two previous studies \citep{weston2014tagspace, zarrella2016mitre} have proven the effectiveness of hashtag generation on stance detection, but have not explored mining external knowledge from a large pre-trained language model via hashtag generation task. 
Given the success of pre-trained language models (PLMs)~\citep{devlin2018bert, liu2019roberta, lewis2019bart, diao2019zen, zhang2019dialogpt, yang2020styledgpt, diao2021taming, diao2022black}, we propose a framework that could leverage the advantages of both generative PLMs and discriminative PLMs for better tweet classification.

\subsection{Generic Keyphrase Generation}
There is a large body of research focused on the keyphrase generation for generic documents, such as news reports \cite{wan2008single}, and scientific documents \cite{meng2017deep}, which mainly consist of two methodology streams, extractive and generative approaches.
In detail, for the keyphrase generation of scientific document,
many previous studies focused on extracting hashtags from documents \citep{hulth2003improved, mihalcea2004textrank, wan2008single, zhang2016keyphrase, luan2017scientific}.
Compared to extractive approaches, generative ones have attracted more attention in recent years for their ability to predict absent keyphrases for an input document.
For example,
\cite{meng2017deep} proposed CopyRNN, which is an early study with attention and copy mechanism.
\cite{chen2018keyphrase} took correlation among multiple keyphrases into consideration 
to eliminate duplicate keyphrases.
To further enhance keyphrase generation, other studies tried to utilize extra information:
\cite{ye2018semi} proposed to assign synthetic keyphrases to unlabeled documents and then use them to enlarge the training data;
\cite{chen2019integrated} retrieved similar documents from the training data for the input document and encoded their keyphrases as external knowledge, while 
\cite{chen2019guided} leveraged title information for this task.
To increase the diversity of keyphrases, a reinforcement learning approach is introduced by \cite{chan2019neural} to encourage their model to generate the correct number of keyphrases with an adaptive reward.
Although existing models are capable of predicting both present and absent keyphrases, there is still potential to facilitate keyphrase generation with unlabeled data such as relevant documents.
In doing so,
\textsc{\ModelName} offers a more effective and efficient solution.

\subsection{Hashtag Generation for Social Media Text}
For social media text, hashtag generation aims to produce hashtags for a given post automatically.
The nature of hashtags makes it difficult for researchers to directly transfer the keyphrase generation methods to this domain mainly because hashtags are short in length and extremely informal. 
Based on the features of social media text, prior work can be categorized into three types: extractive methods, classification methods, and generative methods.
Extractive methods \cite{zhang-etal-2016-keyphrase, zhang2018encoding} try to identify important words from the input post and extract them as hashtags, which suffers from the lack of ability to deal with absent hashtags.
Classification methods \cite{huang2016hashtag, zhang2017hashtag} formulate the hashtag prediction problem from the classification point of view, which aims to classify the true hashtags from a pre-defined candidate list.
However, the hashtags are diverse in social media and a lot of new hashtags are generated by users every day, so we can not include sufficient words in the candidate list.
To produce both present and absent hashtags in a more flexible way, generative methods \cite{wang2019microblog, wang2019topic} try to generate hashtags from scratch following a sequence-to-sequence (Seq2Seq) paradigm, which is capable of generating unseen hashtags in the post without needing a candidate list.
Because the microblog post is short and lacks sufficient information, previous generative methods use either the topic information from the given post or associated conversation information to enrich the Seq2Seq model.
However, only considering topic information or conversation is not enough because they omit the information carried by relevant posts and conversations, and thus the model cannot easily adapt to dynamic semantic context, which is fast changing in social media platforms.
Compared to the above studies, {\ModelName} provides a solution that explicitly incorporates the associated conversation context and the relevant posts and conversations, in a simple architecture.

\section{Conclusion}
This paper proposed a novel two-stage framework for hashtag-guided low-resource tweet classification with a \textbf{Hashtag Generator} and a \textbf{Tweet Classifier}.
The hashtag generator introduces relevant tweets and entities, and summarizes them into hashtags, while the tweet classifier leverages the global contexts to classify the tweets.
One benefit of our model is the capability of leveraging the pre-trained knowledge of both generative models (\textit{e.g.,} BART) and discriminative models (\textit{e.g.,} BERT).
Another benefit is the enriched multi-grained global contexts introduced by the hashtags, especially the absent ones.
To conduct our experiments, we extract the hashtags based on seven tweet classification datasets and use these to train our hashtag models. Meanwhile, for the classification models, we consider a low-resource setting by sampling from the training sets.
Experiments on a tweet classification benchmark demonstrate the effectiveness of our approach in identifying relevant tweets and entities, as well as in generating present and absent hashtags which provide valuable contexts.

\section*{Acknowledgments}
We thank the anonymous reviewers for their valuable suggestions.
This work was supported by the General Research Fund (GRF) of Hong Kong (No. 16310222 and No. 16201320). 
Shizhe Diao was supported by the Hong Kong Ph.D. Fellowship Scheme (HKPFS).

\bibliographystyle{ACM-Reference-Format}
\bibliography{anthology,sample-base}

\clearpage
\appendix
\begin{table*}[h]
    \small
    \centering
    \caption{Examples of how we generate the training data and ground truth for our hashtag seq2seq model. The \textcolor{blue}{blue text} represents the present hashtags, while the \textcolor{red}{red text} represents the absent hashtags.}
    \vspace{-0.2 cm}
    \begin{tabular}{p{190pt}|p{130pt}|p{100pt}}
    \toprule
        \textbf{Original Tweet} & \textbf{Processed Tweet (New Input)}  & \textbf{Hashtags (Target Output)} \\
        \midrule
         Going to \textcolor{blue}{\#BigApple} tomorrow! Loving \textcolor{blue}{\#NJTransit} \textcolor{blue}{\#commute}! See you all! & Going to BigApple tomorrow! Loving NJTransit commute! See you all! & \textcolor{blue}{BigApple, NJTransit, commute} \\
         \hline
         Abortion IS NOT a political issue it is a MORAL issue. \textcolor{red}{\#Catholic \#Christian \#Conservative} & Abortion IS NOT a political issue it is a MORAL issue. &  \textcolor{red}{Catholic, Christian, Conservative} \\
         \hline
         Latest \textcolor{blue}{\#crypto} developments: Top 10 coins to watch \textcolor{red}{\#NotFinancialAdvice} \textcolor{red}{\#DoYourOwnResearch} & Latest crypto developments: Top 10 coins to watch & \textcolor{blue}{crypto}, \textcolor{red}{NotFinancialAdvice}, \textcolor{red}{DoYourOwnResearch} \\
    \bottomrule
    \end{tabular}
    \label{tab:hashtag-dataset}
\end{table*}

\section{Details of Datasets}
\label{appendix: Datasets}
We perform our experiments on 7 diverse tweet classification tasks in the TweetEval benchmark datasets~\citep{barbieri-etal-2020-tweeteval}. 
All 7 tasks are taken from previous SemEval tasks (and corresponding datasets) and are as follows: emotion recognition~\citep{mohammad2018semeval}, emoji prediction \citep{barbieri2018semeval}, irony detection~\citep{van2018semeval}, hate speech detection~\citep{basile-etal-2019-semeval}, offensive language identification~\citep{zampieri2019semeval}, sentiment analysis~\citep{rosenthal2017semeval}, and stance detection~\citep{mohammad2016semeval}. Additional details about these tasks can be found in Table~\ref{tab:dataset_statistics}.

\textbf{Creating the hashtag generation input and output.} For hashtag generation, we train a separate hashtag generator for each task, as the datasets cover significantly different domains. 
To extract hashtags, we searched for appearances of the octothorpe symbol ($\#$) and considered the contiguous string following it, until a whitespace is reached. These hashtags can either appear mid-tweet or at the end of the tweet. For hashtags that appear mid-tweet, we remove the hashtag symbol ($\#$) but keep the word itself, as removing these words may disrupt the coherence of the sentence. On the other hand, for hashtags that appear at the end of the tweet, we completely remove these from the body of the tweet. If we keep these words at the end of the sentence, the hashtag generator will easily pick up on these false signals and just learn to return the last few words of every sentence without actually learning to perform hashtag generation. By considering this setting of keeping in-sentence hashtags and removing end-of-sentence hashtags, we are able to train the model to perform prediction for both present hashtags and absent hashtags. Examples of this overall process are outlined in Table \ref{tab:hashtag-dataset}.
Similar to \citet{wang-etal-2019-microblog}, we implement some preprocessing steps to clean up our tweets. 
Links, mentions, and numbers were replaced with ``URL'', ``MENTION'', and ``DIGIT'', respectively. 
The details of the processed dataset are shown in Table \ref{tab:dataset_statistics}.

\textbf{Low-resource setting.} 
As discussed in Section \ref{sec:intro}, real-life classification labels are often difficult to acquire for the rapidly changing landscape of tweets.
As such, we conduct our experiments in a low-resource setting, where we randomly select a subset from the training set. 
To keep the sizes of different datasets relatively consistent, we use the following sampling ratios: If the size of the original dataset is $<5000$, we sample $10\%$ of the dataset. 
If it's between $5000$ and $10000$, we sample $5\%$, and if it's $>10000$, we sample $1\%$.
All experiments are performed over ten different random seeds. 

\section{Details of Baselines}
\label{appendix: Baselines}
To verify the effectiveness of our {\ModelName} model, we compare its performance against the following baseline classification models: \textbf{Kim-CNN} \cite{kim-2014-convolutional}, \textbf{BiLSTM} \cite{bilstm-1997}, \textbf{BERT}~\citep{devlin2018bert}, \textbf{RoBERTa} \cite{Liu2019RoBERTaAR}, \textbf{BERTweet}~\citep{nguyen-etal-2020-bertweet}, and \textbf{TimeLMs}~\citep{loureiro-etal-2022-timelms}.
BERTweet and TimeLMs are two state-of-the-art models that are pre-trained on 850 million and 124 million English tweets, respectively.

\begin{itemize}[leftmargin=*,label=$\bullet$,noitemsep,partopsep=0pt,topsep=0pt,parsep=0pt]
    \item \textbf{Kim-CNN} \cite{kim-2014-convolutional}. A simple convolutional neural networks framework with a classification layer attached to the end. For this CNN-based model, we use kernel sizes of 2,3,4,5 and 64 filters for each, with a dropout of 0.5 before the linear layer.
    \item \textbf{BiLSTM} \cite{bilstm-1997}. 
    A bidirectional long short-term memory network considering the temporal order of words in the tweet. The hidden size and the dropout rate for the LSTM are set to 512 and 0.2, respectively. We use an LSTM hidden size of 512 and an LSTM dropout of 0.2. In addition, we use a dropout of 0.5 and a size of 32 for the final linear classification layer.
    \item \textbf{BERT}~\citep{devlin2018bert}. A BERT-base model pre-trained on generic corpus with a classification head fine-tuned on specific task.
    \item \textbf{RoBERTa} \cite{Liu2019RoBERTaAR}. 
    A RoBERTa-base model pre-trained on generic corpus with a classification head fine-tuned on specific task. 
    \item \textbf{BERTweet}~\citep{nguyen-etal-2020-bertweet}. A RoBERTa-base model pre-trained on 850M English Tweets with a classification head fine-tuned on specific task.
    \item \textbf{TimeLMs}~\citep{loureiro-etal-2022-timelms}. A RoBERTa-base model specialized on diachronic Twitter data and pre-trained on 124M English Tweets.
    \item \textbf{{\ModelName}} (and its variants) -- For the \textsc{Hash-Gen} hashtag generator, we use the Huggingface BART-base model, while for the tweet classifier, we use the BERTweet and TimeLMs. Additionally, for the document embeddings used in the Tweet Attention Module (TAM), we used the \texttt{all-MiniLM-L6-v2} model from the sentence-transformers \footnote{\url{https://www.sbert.net/}} library.
\end{itemize}

\end{document}